\documentclass{article}
\usepackage{ijcai25}

\usepackage{times}
\usepackage{soul}
\usepackage{url}
\usepackage[hidelinks]{hyperref}
\usepackage[utf8]{inputenc}
\usepackage[small]{caption}
\usepackage{graphicx}
\usepackage{amsmath}
\usepackage{amsthm}
\usepackage{booktabs}
\usepackage{algorithm}
\usepackage{algorithmic}
\usepackage[switch]{lineno}
\usepackage{xcolor}
\usepackage{multirow}
\usepackage{setspace}
\usepackage{tcolorbox} 

\usepackage{threeparttable}

\usepackage{pifont}
\usepackage{fontawesome}

\usepackage{tikz}
\usepackage[edges]{forest}
\definecolor{hiddendraw}{RGB}{205, 44, 36}
\definecolor{hidden-blue}{RGB}{194,232,247}
\definecolor{hidden-orange}{RGB}{243,202,120}
\definecolor{hidden-yellow}{RGB}{242,244,193}

\definecolor{contribution-blue}{HTML}{6C8EBF}
\definecolor{contribution-green}{HTML}{82B366}
\definecolor{contribution-orange}{HTML}{DFAB2B}

\title{Deep Learning for Multivariate Time Series Imputation: A Survey}

\author{
    Author Name
    \affiliations
    Affiliation
    \emails
    email@example.com
}

\author{
Jun Wang$^{1, 2, 6}$\thanks{The first two authors contributed equally to this work.}
\and
Wenjie Du$^{1*}$  \and 
Yiyuan Yang$^{1,3}$  \and
Linglong Qian$^{1,4}$  \and \\
Wei Cao$^1$ \and
Keli Zhang$^5$ \and 
Wenjia Wang$^{2, 6}$ \and
Yuxuan Liang$^6$ \and
Qingsong Wen$^7$\thanks{Corresponding author.}
\vspace{2pt}
\affiliations
$^1$\href{https://pypots.com}{\raisebox{-5px}{\includegraphics[width=20px]{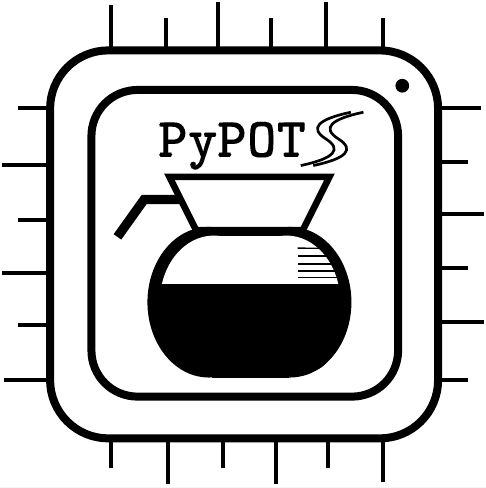}}PyPOTS Research} \quad
$^2$Hong Kong University of Science and Technology  \\
$^3$University of Oxford \quad
$^4$King's College London \quad
$^5$Huawei Noah’s Ark Lab \\
$^6$Hong Kong University of Science and Technology (Guangzhou) \quad
$^7$Squirrel Ai Learning
\vspace{5pt}
\emails
jwangfx@connect.ust.hk,\quad
wdu@time-series.ai,\quad
yiyuan.yang@cs.ox.ac.uk,\quad
linglong.qian@kcl.ac.uk,\quad
weicaomsra@gmail.com,\quad
zhangkeli1@huawei.com,\quad
wenjiawang@ust.hk,\quad
yuxliang@outlook.com,\quad
qingsongedu@gmail.com
}

\begin{document}

\maketitle

\begin{abstract}
	Missing values are ubiquitous in multivariate time series (MTS) data, posing significant challenges for accurate analysis and downstream applications. In recent years, deep learning-based methods have successfully handled missing data by leveraging complex temporal dependencies and learned data distributions. In this survey, we provide a comprehensive summary of deep learning approaches for multivariate time series imputation (MTSI) tasks. We propose a novel taxonomy that categorizes existing methods based on two key perspectives: imputation uncertainty and neural network architecture. Furthermore, we summarize existing MTSI toolkits with a particular emphasis on the PyPOTS Ecosystem, which provides an integrated and standardized foundation for MTSI research. Finally, we discuss key challenges and future research directions, which give insight for further MTSI research. This survey aims to serve as a valuable resource for researchers and practitioners in the field of time series analysis and missing data imputation tasks. A well-maintained MTSI paper and tool list is available at \href{https://github.com/WenjieDu/Awesome_Imputation}{https://github.com/WenjieDu/Awesome\_Imputation}.
\end{abstract}

\section{Introduction}

\begin{figure} [!t]
	\centering
	\includegraphics[width=0.98\columnwidth]{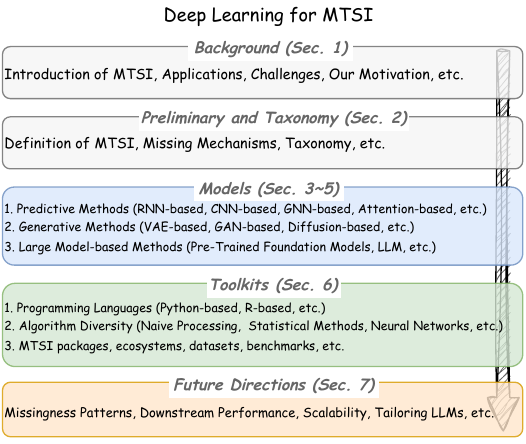}
	\vspace{-10pt}
	\caption{The framework of our survey.}
	\label{fig:framework}
	\vspace{-5pt}
\end{figure}

The data collection process of multivariate time series in various fields is often fraught with difficulties and uncertainty. In IoT systems, sensor failures and unstable environments lead to missing measurements~\cite{li2023data}. Clinical studies face challenges from irregular sampling and privacy concerns~\cite{ibrahim2012missing,esteban2017real,qian2025deep}. Financial and transportation systems encounter data gaps due to system downtime and communication issues~\cite{bai2008forecasting,gong2021missing}. These missing values can significantly affect the accuracy and reliability of downstream analysis and decision-making. In real-world datasets like PhysioNet2012 \cite{silva2012physionet}, missing rates can exceed 80\%. Consequently, exploring how to reasonably and effectively impute missing components in multivariate time series is attractive and essential.

The earlier statistical imputation methods have historically been widely used for handling missing data. 
Those methods substitute the missing values with the statistics (e.g., zero value, mean value, and last observed value~\cite{amiri2016missing}) or simple statistical models, including ARIMA~\cite{bartholomew1971time}, ARFIMA~\cite{hamzaccebi2008improving}, and SARIMA~\cite{hamzaccebi2008improving}.
Furthermore, machine learning techniques like regression, K-nearest neighbor, matrix factorization, etc., have gained prominence in the literature for addressing missing values in multivariate time series. 
Key implementations of these approaches include KNNI~\cite{altman1992introduction}, TIDER~\cite{liu2022multivariate}, MICE~\cite{vanbuuren2011mice}, etc.
While statistical and machine learning imputation methods are simple and efficient, they fall short in capturing the intricate temporal relationships and complex variation patterns inherent in time series data, resulting in limited performance.

More recently, deep learning-based imputation methods have demonstrated strong modelling capabilities for handling missing data. These approaches leverage advanced architectures such as Transformers, Variational Autoencoders (VAEs), Generative Adversarial Networks (GANs), diffusion models, Pre-trained Foundation Models (PFMs), and Large Language Models (LLMs) to capture the underlying structures and complex temporal dynamics of time series data. 
By learning the true data distribution from observed values, deep learning imputation methods can generate more reliable and contextually appropriate estimates for missing components. 
While several surveys exist on imputation techniques~\cite{khayati2020mind,fang2020time}, they primarily focus on statistical and traditional machine learning approaches, offering limited discussion on deep learning-based methods. Given that multivariate time series imputation is a critical \textit{data preprocessing} step for downstream time series analysis, a comprehensive and systematic survey on deep learning-driven imputation methods would provide valuable insights and contribute significantly to the research community.

In this survey paper,  we endeavor to bridge the existing knowledge gap by providing a comprehensive summary of the latest developments in deep learning methods for multivariate time series imputation (MTSI). The framework of this survey is illustrated in Figure~\ref{fig:framework}. In detail, we first present a succinct introduction to the topic, followed by the proposal of a novel taxonomy, categorizing approaches based on two perspectives: \textit{imputation uncertainty} and \textit{neural network architecture}.
Imputation uncertainty quantifies the confidence in estimated values for missing data. To capture this, multiple stochastic samples are generated, and imputations are performed across these variations~\cite{little2019statistical}.
Accordingly, we categorize imputation methods into predictive ones, offering fixed estimates, and generative ones, which provide a distribution of possible values to account for imputation uncertainty.
For neural network architecture, we explore a range of deep learning models tailored for MTSI, including RNN-based ones, GNN-based ones, CNN-based ones, attention-based ones, VAE-based ones, GAN-based ones, diffusion-based ones, PFM-based ones, and LLM-based ones.

To the best of our knowledge, this is the first comprehensive and systematic review of deep learning algorithms in the realm of MTSI, aiming to stimulate further research in this field.
A corresponding resource that has been continuously updated can be found in our GitHub 
repository\footnote{\url{https://github.com/WenjieDu/Awesome\_Imputation}}. In summary, the contributions of this survey include:
\begin{enumerate}
	\item We introduce a novel taxonomy for deep multivariate time series imputation, and categorize them based on imputation uncertainty and neural network architecture.
	\item We provide a comprehensive overview of existing MTSI toolkits. In particular, we highlight the PyPOTS Ecosystem, which integrates diverse imputation algorithms, standardized pipelines, and benchmarking resources, facilitating accessible and reproducible MTSI research.
	\item We identify future directions, including missingness patterns, downstream task integration, and model scalability, offering insights to guide further advancements.
\end{enumerate}

\section{Preliminary and Taxonomy}
\subsection{Background of MTSI}
\paragraph{Problem Definition}
A complete time-series dataset on $[0,T]$ typically can be denoted as $\mathcal{D}=\{\mathbf{X}_{i}, \mathbf{t}_{i}\}_{i=1}^{N}$. 
Hereby, $\mathbf{X}_{i} = \{x_{1:K, 1:L} \} \in \mathcal{R}^{K \times L}$ and $\mathbf{t}_{i} = ({t_{1}, \cdots, t_{L}}) \in [0, T]^{L}$, where $K$ is the number of features and $L$ is the length of time series. 
In the missing data context, each complete time series can be split into an observed and a missing part, i.e., $\mathbf{X}_{i} = \{\mathbf{X}^{o}_{i}, \mathbf{X}^{m}_{i}\}$.
For encoding the missingness, we also denote an observation matrix as $\mathbf{M}_{i} = \{m_{1:K, 1:L} \}$, where $m_{k,l} = 0$ if $x_{k,l}$ is missing at timestamp $t_{l}$, otherwise $m_{k,l} = 1$.
Furthermore, we can also calculate a time-lag matrix $\boldsymbol{\delta}_{i} = \{\delta_{1:K, 1:L} \}$ by the following rule:
\begin{equation}\nonumber
	\delta_{k,l}=\left\{\begin{array}{ll}{0,} & {\text {if } l=1} \\ {t_{l}-t_{l-1},} & {\text {if } m_{k,l-1} =1 \text { and } l>1} \\ {\delta_{k, l-1}+t_{l}-t_{l-1},} & {\text {if } m_{k, l-1}=0 \text { and } l>1}\end{array}\right.
\end{equation}

Hence, each incomplete time series is expressed as $\{\mathbf{X}_{i}^{o}, \mathbf{M}_{i}, \boldsymbol{\delta}_{i}\}$.
The objective of MTSI is to construct an imputation model $\mathcal{M}_{\theta}$, parameterized by $\theta$, to accurately estimate missing values in $\mathbf{X}^{o}$. 
The \emph{imputed} matrix is defined as:
\begin{equation}
	\mathbf{\hat{X}} = \mathbf{{M}} \odot \mathbf{{X}}^{o} + (1 -\mathbf{M}) \odot \mathbf{\bar{X}},
\end{equation}
where $\odot$ denotes element-wise multiplication, and $\mathbf{\bar{X}} = \mathcal{M}_{\theta}(\mathbf{{X}}^{o})$ is the reconstructed matrix. The aim of $\mathcal{M}_{\theta}$ is twofold: (i) to make $\mathbf{\hat{X}}$ approximate the true \emph{complete} data $\mathbf{X}$ as closely as possible, or (ii) to enhance the downstream task performance using $\mathbf{\hat{X}}$ compared to using the original $\mathbf{X}^{o}$.

\paragraph{Missing Mechanism}
The missing mechanisms, i.e., the cause of missing data, define the statistical relationship between observations and the probability of missingness~\cite{nakagawa2015missing}.  
In real-world scenarios, missing mechanisms are inherently complex, and imputation model performance heavily depends on how well our assumptions align with the actual missing data patterns.  
According to Rubin's theory~\cite{rubin1976missing}, missing mechanisms fall into three categories: Missing Completely At Random (MCAR), Missing At Random (MAR), and Missing Not At Random (MNAR). MCAR implies that missingness is independent of both observed and missing data. Conversely, MAR indicates that missingness depends solely on observed data. MNAR suggests that missingness is related to the missing data itself and may also be influenced by observed data.  
These mechanisms can be formally defined as follows:
\begin{itemize}
	\item MCAR: $p(\mathbf{M}|\mathbf{X}) = p(\mathbf{M})$,
	\item  MAR: $p(\mathbf{M}|\mathbf{X}) = p(\mathbf{M}|\mathbf{X}^{o})$,
	\item  MNAR: $p(\mathbf{M}|\mathbf{X}) = p(\mathbf{M}|\mathbf{X}^{o}, \mathbf{X}^{m})$.
\end{itemize}
MCAR and MAR are stronger assumptions compared to MNAR and are considered ``ignorable"~\cite{little2019statistical}. 
This means that the missing mechanism can be disregarded during imputation, focusing solely on learning the data distribution, i.e., $p(\mathbf{X}^{o})$. 
In contrast, MNAR, often more reflective of real-life scenarios, is ``non-ignorable", overlooking its missing mechanism can lead to biased parameter estimates. 
The objective here shifts to learning the joint distribution of the data and its missing mechanism, i.e., $p(\mathbf{X}^{o}, \mathbf{M})$.

\subsection{Taxonomy of Deep Learning-based MTSI}

\begin{figure} [!ht]
	\centering
	\includegraphics[width=1\columnwidth]{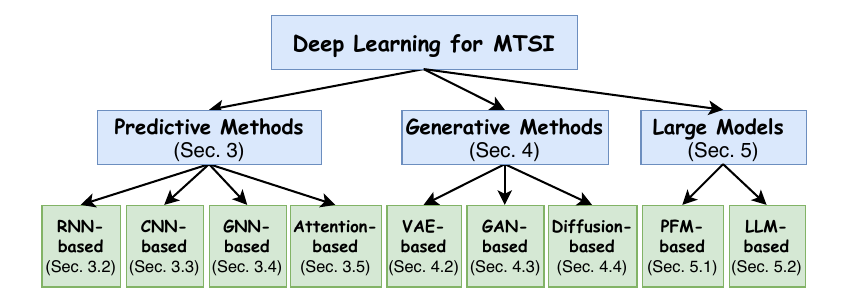} 
	\caption{The taxonomy of deep learning methods for multivariate time series imputation from the view of imputation uncertainty and neural network architecture.}
	\label{fig:taxonomy}
\end{figure}

\begin{table*}[!ht]
	\centering
	\begin{spacing}{1}
		\resizebox{1\linewidth}{!}{
			\renewcommand{\arraystretch}{1.05}{
				\begin{tabular}{l|llcll}
					\toprule
					\textbf{Method}  &    \textbf{Venue}&   \textbf{Category}     &    \textbf{Imputation Uncertainty} & \textbf{Neural Network Architecture}   &  \textbf{Missing Mechanism}\\
					\hline
					GRU-D~\cite{che2018grud} & Scientific Reports & predictive & \faTimes   &  RNN &MCAR  \\ 
					M-RNN~\cite{yoon2017mrnn} & TBME & predictive & \faTimes & RNN  & MCAR \\  
					BRITS~\cite{cao2018brits}  & NeurIPS & predictive & \faTimes & RNN  &  MCAR\\  
					TimesNet~\cite{wu2023timesnet} & ICLR & predictive & \faTimes & CNN  & MCAR \\  
					GRIN~\cite{cini2022grin} & ICLR & predictive & \faTimes & GNN  & MCAR / MAR \\  
					SPIN~\cite{marisca2022learning} & NeurIPS & predictive & \faTimes & GNN, Attention  & MCAR / MAR \\  
					Transformer~\cite{vaswani2017attention}& NeurIPS  & predictive & \faTimes & Attention  & MCAR\\ 
					SAITS~\cite{du2023saits} & ESWA  &predictive & \faTimes &  Attention  &  MCAR\\ 
					DeepMVI~\cite{bansal2021deepmvi} & VLDB & predictive & \faTimes & Attention, CNN  &  MCAR \\ 
					ImputeFormer~\cite{nie2024imputeformer} &KDD & predictive & \faTimes & Attention & MCAR \\
					Casper~\cite{jing2024causality} &CIKM & predictive & \faTimes & GNN, Attention & MCAR \\
					HSPGNN~\cite{liang2024higher} &CIKM & predictive & \faTimes & GNN & MCAR \\
					GP-VAE~\cite{fortuin2020gpvae}& AISTATS & generative & \faCheckCircleO & VAE, CNN   &  MCAR / MAR \\ 
					V-RIN~\cite{mulyadi2021uncertainty}& Trans. Cybern. & generative & \faCheck & VAE, RNN  & MCAR / MAR\\
					supnotMIWAE~\cite{kim2023probabilistic}& ICML & generative & \faCheckCircleO & VAE  &MNAR \\ 
					GRUI-GAN~\cite{luo2018grui} &NeurIPS & generative & \faCheckCircleO & GAN, RNN  & MCAR\\ 
					E$^2$GAN~\cite{luo2019e2gan}&IJCAI & generative & \faCheckCircleO & GAN, RNN  & MCAR \\ 
					NAOMI~\cite{liu2019naomi} &NeurIPS & generative & \faCheckCircleO & GAN, RNN  & MCAR\\ 
					SSGAN~\cite{miao2021ssgan} &AAAI  & generative & \faCheckCircleO & GAN, RNN    & MCAR \\ 
					CSDI~\cite{tashiro2021csdi}& NeurIPS & generative & \faCheckCircleO & Diffusion, Attention, CNN  & MCAR \\
					SSSD~\cite{alcaraz2023sssd}& TMLR & generative & \faCheckCircleO & Diffusion, Attention  & MCAR \\
					CSBI~\cite{chen2023csbi} &ICML & generative & \faCheckCircleO & Diffusion, Attention  & MCAR\\
					MIDM~\cite{wang2023observed} &KDD & generative & \faCheckCircleO & Diffusion, Attention  & MCAR\\
					PriSTI~\cite{liu2023pristi} &ICDE & generative & \faCheckCircleO & Diffusion, Attention, GNN, CNN  & MCAR \\
					SPD~\cite{bilovs2023dspd} &ICML & generative & \faCheckCircleO & Diffusion, Attention & MCAR \\
					SADI~\cite{dai2024sadi} &AISTATS & generative & \faCheck & Diffusion, Attention & MCAR / MAR / MNAR \\
					FGTI~\cite{yang2024frequency} &NeurIPS & generative & \faCheck & Diffusion, Attention & MCAR / MAR / MNAR  \\
					MTSCI~\cite{zhou2024mtsci} &CIKM & generative & \faCheck & Diffusion, Attention & MCAR \\
					MOMENT~\cite{goswami2024moment} & ICML & large model & \faCheck  & Foundation model & MCAR \\
					Timer~\cite{liutimer} & ICML & large model & \faCheck  & Foundation model & MCAR \\
					Timemixer++~\cite{wang2024timemixer++}& ICLR & large model & \faCheck  & Foundation model & MCAR \\
					GPT4TS~\cite{zhou2023one} & NeurIPS & large model & \faCheck  & Large language model & MCAR \\
					LLM-TS Integrator~\cite{chen2024enhance}& NeurIPS workshop & large model & \faCheck  & Large language model & MCAR \\
					\bottomrule
			\end{tabular}}
		}
		\caption{Summary of deep learning methods for multivariate time series imputation. \faCheck ~and \faCheckCircleO ~indicate methods capable of accounting for imputation uncertainty, whereas \faTimes~denotes methods that do not. Furthermore, \faCheck ~denotes that the methods also define the fidelity score to explicitly measure the imputation uncertainty.}
		\label{tab:taxonomy}
	\end{spacing}
	\vspace{-3mm}
\end{table*}

To summarize the existing deep multivariate time series imputation methods, we propose a taxonomy from the perspectives of {imputation uncertainty} and neural network architecture as illustrated in Figure~\ref{fig:taxonomy}, and provide a more detailed summary of these methods in Table~\ref{tab:taxonomy}.
Regarding large model-based methods, we separate them as a category considering their application strategies are quite different from general neural network algorithms.
{For imputation uncertainty, we categorize imputation methods into predictive and generative types, based on their ability to yield varied imputations that reflect the inherent uncertainty in the imputation process.}
In the context of the neural network architecture, we examine prominent deep learning models specifically designed for multivariate time series imputation. 
The discussed models encompass  RNN-based ones, CNN-based ones, GNN-based ones, attention-based ones, VAE-based ones, GAN-based ones, and diffusion-based ones.
In the following sections, we will delve into and discuss the existing deep time series imputation methods from these two perspectives.

\section{Predictive Methods}
This section delves into predictive imputation methods, and our discussion primarily focuses on four types: RNN-based, CNN-based, GNN-based, and attention-based models.

\subsection{Learning Objective}
{Predictive imputation methods consistently predict deterministic values for the same missing components, thereby not accounting for the uncertainty in the imputed values.}
Typically, these methods employ a reconstruction-based learning manner with the learning objective being,
\begin{equation}
	\mathcal{L}_{det}(\theta) = \sum_{i=1}^{N}  \ell_{e}(\mathbf{M}_{i} \odot {\mathbf{\bar{X}}_{i}}, \mathbf{M}_{i} \odot \mathbf{X}^{o}_{i}),
\end{equation}

\noindent where $\ell_{e}$ is an absolute or squared error function.

\subsection{RNN-based Models}
As a natural way to model sequential data, Recurrent Neural Networks (RNNs) were developed early on the topic of advanced time-series analysis, and imputation is not an exception.
GRU-D~\cite{che2018grud}, a variant of GRU, is designed to process time series containing missing values. 
It is regulated by a temporal decay mechanism, which takes the time-lag matrix $\mathbf{\delta}_{i}$ as input and models the temporal irregularity caused by missing values.
Temporal belief memory~\cite{kim2018tbm}, inspired by a biological neural model called the Hodgkin–Huxley model, is proposed to handle missing data by computing a belief of each feature's last observation with a bidirectional RNN and imputing a missing value based on its corresponding belief. 
M-RNN~\cite{yoon2017mrnn} is an RNN variant that works in a multi-directional way. This model interpolates within data streams with a bidirectional RNN model and imputes across data streams with a fully connected network. 
BRITS~\cite{cao2018brits} models incomplete time series with a bidirectional RNN. It takes missing values as variables of the RNN graph and fills in missing data with the hidden states from the RNN. In addition to imputation, BRITS is capable of working on the time series classification task simultaneously. 
Both M-RNN and BRITS adopt the temporal decay function from GRU-D to capture the informative missingness for performance improvement. 
Subsequent works, such as~\cite{luo2018grui,luo2019e2gan,liu2019naomi,miao2021ssgan}, combine RNNs with the GAN structure to output imputation with higher accuracy. 

\subsection{CNN-based Models}
Convolutional Neural Networks (CNNs) represent a foundational deep learning architecture, extensively employed in sophisticated time series analysis. 
TimesNet~\cite{wu2023timesnet} innovatively incorporates Fast Fourier Transform to restructure 1D time series into a 2D format, facilitating the utilization of CNNs for data processing.
Also in GP-VAE~\cite{fortuin2020gpvae}, CNNs play the role of the backbone in both the encoder and decoder.
Furthermore, CNNs serve as pivotal feature extractors within attention-based models like DeepMVI~\cite{bansal2021deepmvi}, as well as in diffusion-based models such as CSDI~\cite{tashiro2021csdi}, by mapping input data into an embedding space for subsequent processing.

\subsection{GNN-based Models}
GNN-based models treat time series as graph sequences, reconstructing missing values via learned node representations.  
\cite{cini2022grin} introduces GRIN, the first graph-based recurrent architecture for MTSI, leveraging a bidirectional graph recurrent neural network to capture temporal dynamics and spatial similarities, significantly improving imputation accuracy.  
SPIN~\cite{marisca2022learning} further integrates a sparse spatiotemporal attention mechanism into the GNN framework, mitigating GRIN’s error propagation and enhancing robustness against data sparsity.

\subsection{Attention-based Models}
Since Transformer is proposed in~\cite{vaswani2017attention}, the self-attention mechanism has been widely used to model sequence data including time series~\cite{wen2023transformers}.
DeepMVI~\cite{bansal2021deepmvi} integrates transformers with convolutional techniques, tailoring key-query designs to effectively address missing value imputation. 
For each time series, DeepMVI harnesses attention mechanisms to concurrently distill long-term seasonal, granular local, and cross-dimensional embeddings, which are concatenated to predict the final output. 
SAITS~\cite{du2023saits} employs a self-supervised training scheme to deal with missing data, which integrates dual joint learning tasks: a masked imputation task and an observed reconstruction task.
This method, featuring two diagonal-masked self-attention blocks and a weighted-combination block, leverages attention weights and missingness indicators to enhance imputation precision. Besides, ImputeFormer~\cite{nie2024imputeformer} introduces a novel Transformer-based framework that leverages self-attention and temporal context modeling to accurately recover missing values.
In addition to the above models, the attention mechanism is also widely adapted to build the denoising network in diffusion models like CSDI~\cite{tashiro2021csdi}, MIDM~\cite{wang2023observed}, PriSTI~\cite{liu2023pristi}, Diffusion-TS~\cite{yuan2024diffusion} and in GNN-based models like Casper~\cite{jing2024causality} and HSPGNN~\cite{liang2024higher}. 

\subsection{Pros and Cons}
This subsection synthesizes the strengths and challenges of the predictive imputation methods discussed. 
RNN-based models, while adept at capturing sequential information, are inherently limited by their sequential processing nature and memory constraints, which may lead to scalability issues with long sequences~\cite{khayati2020mind}. 
Although CNNs have decades of development and are useful feature extractors to capture neighborhood information and local connectivity, their kernel size and working mechanism intrinsically limit their performance on time-series data as the backbone.
Due to the attention mechanism, attention-based models generally outperform RNN-based and CNN-based methods in imputation tasks due to their superior ability to handle long-range dependencies and parallel processing capabilities. 
GNN-based methods provide a deeper understanding of spatio-temporal dynamics, yet they often come with increased computational complexity, posing challenges for large-scale or high-dimensional data. 

\section{Generative Methods}
In this section, we examine generative imputation methods, including three primary types: VAE-based, GAN-based, and diffusion-based models. 

\subsection{Learning Objective}
{Generative methods are essentially built upon generative models like VAEs, GANs, and diffusion models. 
	They are characterized by their ability to generate varied outputs for missing observations, enabling the quantification of imputation uncertainty.
	Typically, these methods learn probability distributions from the observed data and subsequently generate slightly different values aligned with these learned distributions for the missing observation.}
The primary learning objective of generative methods is thus defined as,
\begin{equation}
	\mathcal{L}_{pro}(\theta) = \sum_{i=1}^{N}  \log p_{\theta}(\mathbf{{X}}^{o}_{i}).
\end{equation}
\noindent where $\theta$ is the model parameters of the imputation model $\mathcal{M}$.

\subsection{VAE-based Models}
VAEs employ an encoder-decoder structure to approximate the true data distribution by maximizing the Evidence Lower Bound (ELBO) on the marginal likelihood. 
This ELBO enforces a Gaussian-distributed latent space from which the decoder reconstructs diverse data points.

The authors in~\cite{fortuin2020gpvae} propose the first VAE-based imputation method GP-VAE, where they utilized a Gaussian process prior in the latent space to capture temporal dynamics.
Moreover, 
the ELBO in GP-VAE is only evaluated on the observed features of the data.
{Authors in~\cite{mulyadi2021uncertainty} design V-RIN to mitigate the risk of biased estimates in missing value imputation. 
	V-RIN captures uncertainty by accommodating a Gaussian distribution over the model output, specifically interpreting the variance of the reconstructed data from a VAE model as an uncertainty measure. 
	It then models temporal dynamics and seamlessly integrates this uncertainty into the imputed data through an uncertainty-aware GRU.}
More recently, authors in ~\cite{kim2023probabilistic} propose supnotMIWAE and introduce an extra classifier, where they extend the ELBO in GP-VAE to model the joint distribution of the observed data, its mask matrix, and its label.
In this way, their ELBO effectively models the imputation uncertainty, and the additional classifier encourages the VAE model to produce missing values that are more advantageous for the downstream classification task.

\subsection{GAN-based Models} \label{gan-based-models}
GANs facilitate adversarial training through a minimax game between two components: a generator aiming to mimic the real data distribution, and a discriminator tasked with distinguishing between the generated and real data. 
This dynamic fosters a progressive refinement of synthetic data that increasingly resembles real samples.

In~\cite{luo2018grui}, authors propose a two-stage GAN imputation method (GRUI-GAN), which is the first GAN-based method for imputing time-series data.
GRUI-GAN first learns the distribution of the observed multivariate time-series data by a standard adversarial training manner and then optimizes the input noise of the generator to further maximize the similarity of the generated and observed multivariate time series.
However, the second stage in GRUI-GAN needs a lot of time to find the best matched input vector, and this vector is not always the best especially when the initial value of the “noise” is not properly set.
Then, an end-to-end GAN imputation model $E^2$GAN~\cite{luo2019e2gan} is further proposed, where the generator takes a denoising autoencoder module to avoid the “noise” optimization stage in GRUI-GAN.
Meanwhile, authors in ~\cite{liu2019naomi} propose a non-autoregressive multi-resolution GAN model (NAOMI), where the generator is assembled by a forward-backward encoder and a multiresolution decoder.
The imputed data are recursively generated by the multiresolution decoder in a non-autoregressive manner, which mitigates error accumulation in scenarios involving high-missing and long sequence time series. 
On the other hand, in~\cite{miao2021ssgan}, authors propose USGAN, which generates high-quality imputed data by integrating a discriminator with a temporal reminder matrix.
This matrix introduces added complexity to the training of the discriminator and subsequently leads to improvements in the generator's performance.
Furthermore, they extend USGAN to a semi-supervised model SSGAN, by introducing an extra classifier.
In this way, SSGAN leverages label information, allowing the generator to estimate missing values while conditioning on observed components and labels simultaneously.

\subsection{Diffusion-based Models}
As an emerging and potent category of generative models, diffusion models are adept at capturing complex data distributions by progressively adding and then reversing noise through a Markov chain of diffusion steps. 
Distinct from VAE, these models utilize a fixed training procedure and operate with high-dimensional latent variables that retain the dimensionality of the input data~\cite{yang2024survey}.

CSDI, introduced in~\cite{tashiro2021csdi}, stands out as the pioneering diffusion model specifically designed for MTSI. 
Different from conventional diffusion models, CSDI adopts a conditioned training approach, where a subset of observed data is utilized as conditional information to facilitate the generation of the remaining segment of observed data.
However, the denoising network in CSDI relies on two transformers, exhibiting quadratic complexity concerning the number of variables and the time series length. 
This design limitation raises concerns about memory constraints, particularly when modeling extensive multivariate time series.
In response to this challenge, a subsequent work by~\cite{alcaraz2023sssd} introduces SSSD, which addresses the quadratic complexity issue by replacing transformers with structured state space models~\cite{gu2022efficiently}. 
This modification proves advantageous, especially when handling lengthy multivariate time series, as it mitigates the risk of memory overflow.
Another approach CSBI, introduced in ~\cite{chen2023csbi}, improves the efficiency by modeling the diffusion process as a Schrodinger bridge problem, which could be transformed into computation-friendly stochastic differential equations. Also, SADI~\cite{dai2024sadi} is a similarity-aware diffusion model that leverages a self-attention mechanism to capture inter-patient similarities for effective imputation of missing values. MTSCI~\cite{zhou2024mtsci} leverages cross-channel correlations and multi-scale temporal dynamics features to effectively recover missing values.

Moreover, the efficacy of diffusion models is notably influenced by the construction and utilization of conditional information. 
MIDM~\cite{wang2023observed} proposes to sample noise from a distribution conditional on observed data's representations in the denoising process, 
In this way, it can explicitly preserve the intrinsic correlations between observed and missing data. 
PriSTI~\cite{liu2023pristi} introduces the spatiotemporal dependencies as conditional information, i.e., provides the denoising network with spatiotemporal attention weights calculated by the conditional feature for spatiotemporal imputation. 
Besides, FGTI~\cite{yang2024frequency} is a frequency-guided framework that leverages spectral analysis to effectively capture both global periodic patterns and local temporal dynamics, thereby enhancing missing data recovery.

Contrasting with the above diffusion-based methods that treat time series as discrete time steps, SPD~\cite{bilovs2023dspd} views time series as discrete realizations of an underlying continuous function and generates data for imputation using stochastic process diffusion.
In this way, SPD posits the continuous noise process as an inductive bias for the irregular
time series, so as to better capture the true generative process, especially with the inherent stochasticity of the data.

\subsection{Pros and Cons}
This subsection delineates the advantages and limitations of the aforementioned generative imputation models.
VAE-based models are adept at modeling probabilities explicitly and offering a theoretical foundation for understanding data distributions. 
However, they are often constrained by their generative capacity, which can limit their performance in capturing complex data variability. 
GAN-based models, on the other hand, excel in data generation, providing high-quality imputations with impressive fidelity to the original data distributions. 
Yet, they are notoriously challenging to train due to issues like vanishing gradients \cite{wu2023differentiable}, which can hamper model stability and convergence. 
Diffusion-based models emerge as powerful generative tools with a strong capacity for capturing intricate data patterns. 
Nevertheless, their computational complexity is considerable, and they also suffer from issues related to boundary coherence between missing and observed parts~\cite{Lugmayr2022repaint}.

\section{Large Model-based Methods}
Large models aim to tackle three critical challenges in MTSI tasks: complex temporal dependencies across multiple scales, diverse missingness patterns, and the need for robust generalization with limited domain data. In this section, we examine how pre-trained foundation models (PFMs) and large language models (LLMs) approach these challenges through distinct but complementary paths.

\subsection{Pre-Trained Foundation Models}
PFMs leverage large-scale pretraining on diverse datasets to enhance generalization and adaptability~\cite{liang2024foundation}. For example, by pretraining on a vast collection of multivariate time series, Timer~\cite{liutimer} learns a rich set of temporal representations, enabling it to generalize across different domains. It employs a self-supervised contrastive learning framework that refines its ability to reconstruct missing values while maintaining consistency in long-range dependencies.
Building on the idea of large-scale pretraining, MOMENT~\cite{goswami2024moment} introduces the Time Series Pile and facilitates pretraining of high-capacity Transformer models using masked time series prediction tasks. 
On the other hand, NuwaTS~\cite{chengnuwats} repurposes pre-trained language models for time series imputation, utilizing specialized embeddings and contrastive learning to handle various missing data patterns across domains.
Meanwhile, Timemixer++~\cite{wang2024timemixer++} explores an alternative to Transformer-based architectures by adopting token-mixing techniques. By mixing temporal and feature-wise representations, Timemixer++ effectively captures both short- and long-term dependencies while reducing computational overhead. 
These PFMs represent a significant step forward in MTSI, offering scalable, generalizable, and high-performance solutions that adapt to various missing data patterns across domains.

\subsection{Large Language Models}
LLMs have shown good capabilities in sequential modeling due to their autoregressive nature, which enables them to capture sequence dependencies through next-token prediction. Their extensive model parameters equip them with a strong learning capacity, making them suitable for handling MTSI tasks~\cite{jin2023large}.
For example, GPT4TS~\cite{zhou2023one} optimizes GPT-2 for MTSI by introducing a key architectural adjustment—the freezing of the attention module. This design allows the model to focus on fine-tuning positional embeddings and layer normalization layers using a small number of samples. Such targeted fine-tuning significantly improves the model’s ability to reconstruct missing values with minimal data availability. Furthermore, LLM-TS Integrator~\cite{chen2024enhance} enhances LLM-based MTSI tasks by integrating statistical and deep learning methods within a hybrid framework. It introduces an adaptive retrieval mechanism to select relevant historical patterns and a self-correction module for iterative refinement. By leveraging retrieval-augmented generation, the model effectively handles irregular and sparse time-series data.

These large model approaches represent a fundamental shift in time series imputation: from treating missing values as isolated points to understanding them within a rich temporal and cross-variable context. While showing promising results, particularly in scenarios with complex missingness patterns, their adoption requires careful consideration of computational resources and domain-specific constraints. 

\section{Time Series Imputation Toolkits}

On the MTSI task, there are existing libraries providing naive processing ways, statistical methods, machine learning imputation algorithms, and deep learning imputation neural networks for convenient usage. 

\textbf{\texttt{imputeTS}}~\cite{moritz2017imputets}, a library in R provides several naive approaches (e.g., mean values, last observation carried forward, etc.) and commonly-used imputation algorithms (e.g., linear interpolation, Kalman smoothing, and weighted moving average) but only for univariate time series. 
Another well-known R package, \textbf{\texttt{mice}}~\cite{vanbuuren2011mice}, implements the method called multivariate imputation by chained equations to tackle missingness in data. 
Although it is not for time series specifically, it is widely used in practice for multivariate time-series imputation, especially in the field of statistics. 
\textbf{\texttt{Impyute}}\footnote{\url{https://github.com/eltonlaw/impyute}} and \textbf{\texttt{Autoimpute}}\footnote{\url{https://github.com/kearnz/autoimpute}} both offer naive imputation methods for cross-sectional data and time-series data. 
\textbf{\texttt{Impyute}} is only with simple approaches like the moving average window, and \textbf{\texttt{Autoimpute}} integrates parametric methods, for example, polynomial interpolation and spline interpolation. 
More recently, \textbf{\texttt{GluonTS}}~\cite{alexandrov2020gluonts}, a generative machine-learning package for time series, provides some naive ways, such as dummy value imputation and casual mean value imputation, to handle missing values. 
In addition to simple and non-parametric methods, \textbf{\texttt{Sktime}}~\cite{loning2019sktime} implements one more option that allows users to leverage integrated machine learning imputation algorithms to fit and predict missing values in the given data, though this works in a univariate way.
\textbf{\texttt{ImputeBench}}~\cite{khayati2020mind} offers a collection of machine learning and deep learning-based imputation but lacks uniform programming languages.

When it comes to deep-learning imputation, \textbf{\texttt{PyPOTS}}~\cite{du2023pypots} is a toolbox focusing on modeling partially-observed time series end-to-end. It contains dozens of neural networks for tasks on incomplete time series, including 37 imputation models so far. Leveraging the toolkits from PyPOTS Ecosystem\footnote{\url{https://pypots.com/ecosystem}} that we developed, 
\textbf{TSI-Bench Suite}~\cite{du2024tsi} provides a set of standard pipelines processing 172 public time-series datasets for benchmarking time-series imputation algorithms. Our comprehensive benchmark results from 34,804 experiments including 28 algorithms, 8 typical datasets from different domains, and diverse missingness patterns are presented in~\cite{du2024tsi}.

\section{Future Direction} \label{future}

\paragraph{Missingness Patterns}
Existing imputation algorithms predominantly operate under the MCAR or MAR assumptions.
However, real-world missing data mechanisms are often more complex, with MNAR being prevalent across various domains. 
The non-ignorable nature of MNAR indicates a fundamental distributional shift between observed and true data. 
For example, in airflow signal analysis, the absence of high-value observations leads to saturated peaks, visibly skewing the observed data distribution compared to the true underlying one.
This scenario illustrates how imputation methods may incur inductive bias in model parameter estimation and underperform in the presence of MNAR.
Addressing missing data in MNAR contexts, distinct from MCAR and MAR, calls for innovative methodologies to achieve better performance.

\paragraph{Downstream Performance}
The primary objective of imputing missing values lies in enhancing downstream data analytics, particularly in scenarios with incomplete information. The prevalent approach is the ``\textit{impute and predict}" \textbf{two-stage paradigm}, where missing value imputation is a part of data preprocessing and followed by task-specific downstream models (e.g. a classifier), either in tandem or sequentially. An alternative method is the ``\textit{encode and predict}" \textbf{end-to-end paradigm}, encoding the incomplete data into a proper representation for multitask learning, including imputation and other tasks (e.g. classification and forecasting, etc.).
{Despite the optimal paradigm for partially-observed time series still remains an open area for future investigation, the latter end-to-end way turns out to be more promising especially when information embedded in the missing patterns is helpful to the downstream tasks~\cite{miyaguchi2021variational}.}

\paragraph{Scalability}
While deep learning imputation algorithms have shown impressive performance, their computational cost often exceeds that of statistical and machine learning-based counterparts. 
In the era of burgeoning digital data, spurred by advancements in communication and IoT devices, we are witnessing an exponential increase in data generation. 
This surge, accompanied by the prevalence of incomplete datasets, poses significant challenges in training deep models effectively~\cite{wu2023differentiable}. 
Specifically, the high computational demands of existing deep imputation algorithms render them less feasible for large-scale datasets. 
Consequently, there is a growing need for scalable deep imputation solutions, leveraging parallel and distributed computing techniques, to effectively address the challenges of large-scale missing data.

\paragraph{Large Language Models for MTSI}
While large models have shown promising results in time series imputation, several critical research directions remain unexplored. First, beyond current architectural innovations, the explicit incorporation of domain-specific temporal constraints and prior knowledge about missingness mechanisms into the pretraining process offers significant potential. Second, while the latest models have advanced temporal modeling, there remains room for fundamental innovations in processing irregular temporal patterns, particularly through more efficient and interpretable architectures. Third, the potential of multimodal learning deserves further investigation, where large models' ability to process different data modalities could incorporate auxiliary information (such as textual descriptions or metadata) to achieve more accurate and contextually appropriate imputations. These directions, coupled with robust evaluation frameworks assessing uncertainty quantification and temporal consistency, could significantly advance time series imputation in critical applications.

\section{Conclusion}

This survey presents a systematic review of deep learning-based methods for multivariate time series imputation with a novel taxonomy to categorize predictive and generative methods and also discusses the large model for MTSI tasks. We provide a comprehensive architecture overview, highlighting their strengths, limitations, and applications. To advance this field, we identify key challenges, including handling MNAR missingness, integrating imputation with downstream tasks, and improving scalability. Future research should explore large-scale pre-trained models and multimodal learning to enhance robustness and real-world applicability.

\small
\bibliographystyle{named}
\bibliography{ijcai25}

\begin{thebibliography}{}

\bibitem[\protect\citeauthoryear{Alcaraz and
  Strodthoff}{2023}]{alcaraz2023sssd}
Juan~Lopez Alcaraz and Nils Strodthoff.
\newblock Diffusion-based time series imputation and forecasting with
  structured state space models.
\newblock {\em Transactions on Machine Learning Research}, 2023.

\bibitem[\protect\citeauthoryear{Alexandrov \bgroup \em et al.\egroup
  }{2020}]{alexandrov2020gluonts}
Alexander Alexandrov, Konstantinos Benidis, Michael Bohlke-Schneider, Valentin
  Flunkert, Jan Gasthaus, et~al.
\newblock {GluonTS: Probabilistic and Neural Time Series Modeling in Python}.
\newblock {\em Journal of Machine Learning Research}, 21(116):1--6, 2020.

\bibitem[\protect\citeauthoryear{Altman}{1992}]{altman1992introduction}
Naomi~S Altman.
\newblock An introduction to kernel and nearest-neighbor nonparametric
  regression.
\newblock {\em The American Statistician}, 46(3):175--185, 1992.

\bibitem[\protect\citeauthoryear{Amiri and Jensen}{2016}]{amiri2016missing}
Mehran Amiri and Richard Jensen.
\newblock Missing data imputation using fuzzy-rough methods.
\newblock {\em Neurocomputing}, 205(1):152--164, 2016.

\bibitem[\protect\citeauthoryear{Bai and Ng}{2008}]{bai2008forecasting}
Jushan Bai and Serena Ng.
\newblock Forecasting economic time series using targeted predictors.
\newblock {\em Journal of Econometrics}, 146(2):304--317, 2008.

\bibitem[\protect\citeauthoryear{Bansal \bgroup \em et al.\egroup
  }{2021}]{bansal2021deepmvi}
Parikshit Bansal, Prathamesh Deshpande, and Sunita Sarawagi.
\newblock Missing value imputation on multidimensional time series.
\newblock In {\em VLDB}, 2021.

\bibitem[\protect\citeauthoryear{Bartholomew}{1971}]{bartholomew1971time}
David~J Bartholomew.
\newblock Time series analysis forecasting and control.
\newblock {\em Journal of the Operational Research Society}, 22(2):199--201,
  1971.

\bibitem[\protect\citeauthoryear{Bilo\v{s} \bgroup \em et al.\egroup
  }{2023}]{bilovs2023dspd}
Marin Bilo\v{s}, Kashif Rasul, Anderson Schneider, Yuriy Nevmyvaka, and Stephan
  G\"{u}nnemann.
\newblock Modeling temporal data as continuous functions with stochastic
  process diffusion.
\newblock In {\em ICML}, 2023.

\bibitem[\protect\citeauthoryear{Cao \bgroup \em et al.\egroup
  }{2018}]{cao2018brits}
Wei Cao, Dong Wang, Jian Li, Hao Zhou, Lei Li, and Yitan Li.
\newblock Brits: Bidirectional recurrent imputation for time series.
\newblock {\em NeurIPS}, 2018.

\bibitem[\protect\citeauthoryear{Che \bgroup \em et al.\egroup
  }{2018}]{che2018grud}
Zhengping Che, Sanjay Purushotham, Kyunghyun Cho, David Sontag, and Yan Liu.
\newblock Recurrent neural networks for multivariate time series with missing
  values.
\newblock {\em Scientific Reports}, 8(1), Apr 2018.

\bibitem[\protect\citeauthoryear{Chen \bgroup \em et al.\egroup
  }{2023}]{chen2023csbi}
Yu~Chen, Wei Deng, Shikai Fang, Fengpei Li, Nicole~Tianjiao Yang, Yikai Zhang,
  et~al.
\newblock Provably convergent schrödinger bridge with applications to
  probabilistic time series imputation.
\newblock In {\em ICML}, 2023.

\bibitem[\protect\citeauthoryear{Chen \bgroup \em et al.\egroup
  }{2024}]{chen2024enhance}
Can Chen, Gabriel~L Oliveira, Hossein Sharifi-Noghabi, and Tristan Sylvain.
\newblock Enhance time series modeling by integrating llm.
\newblock In {\em NeurIPS Workshop on Time Series in the Age of Large Models},
  2024.

\bibitem[\protect\citeauthoryear{Cheng \bgroup \em et al.\egroup
  }{}]{chengnuwats}
Jinguo Cheng, Chunwei Yang, Wanlin Cai, Yuxuan Liang, Qingsong Wen, and Yuankai
  Wu.
\newblock Nuwats: a foundation model mending every incomplete time series.

\bibitem[\protect\citeauthoryear{Cini \bgroup \em et al.\egroup
  }{2022}]{cini2022grin}
Andrea Cini, Ivan Marisca, and Cesare Alippi.
\newblock Filling the g\_ap\_s: Multivariate time series imputation by graph
  neural networks.
\newblock In {\em ICLR}, 2022.

\bibitem[\protect\citeauthoryear{Dai \bgroup \em et al.\egroup
  }{2024}]{dai2024sadi}
Zongyu Dai, Emily Getzen, and Qi~Long.
\newblock Sadi: Similarity-aware diffusion model-based imputation for
  incomplete temporal ehr data.
\newblock In {\em International Conference on Artificial Intelligence and
  Statistics}, pages 4195--4203. PMLR, 2024.

\bibitem[\protect\citeauthoryear{Du \bgroup \em et al.\egroup
  }{2023}]{du2023saits}
Wenjie Du, David Cote, and Yan Liu.
\newblock {SAITS: Self-Attention-based Imputation for Time Series}.
\newblock {\em Expert Systems with Applications}, 219:119619, 2023.

\bibitem[\protect\citeauthoryear{Du \bgroup \em et al.\egroup
  }{2024}]{du2024tsi}
Wenjie Du, Jun Wang, Linglong Qian, Yiyuan Yang, Zina Ibrahim, Fanxing Liu,
  Zepu Wang, Haoxin Liu, Zhiyuan Zhao, Yingjie Zhou, et~al.
\newblock Tsi-bench: Benchmarking time series imputation.
\newblock {\em arXiv preprint arXiv:2406.12747}, 2024.

\bibitem[\protect\citeauthoryear{Du}{2023}]{du2023pypots}
Wenjie Du.
\newblock {PyPOTS: a Python toolbox for data mining on Partially-Observed Time
  Series}.
\newblock In {\em SIGKDD workshop on Mining and Learning from Time Series},
  2023.

\bibitem[\protect\citeauthoryear{Esteban \bgroup \em et al.\egroup
  }{2017}]{esteban2017real}
Crist{\'o}bal Esteban, Stephanie~L Hyland, and Gunnar R{\"a}tsch.
\newblock Real-valued (medical) time series generation with recurrent
  conditional gans.
\newblock {\em ArXiv Preprint ArXiv:1706.02633}, 2017.

\bibitem[\protect\citeauthoryear{Fang and Wang}{2020}]{fang2020time}
Chenguang Fang and Chen Wang.
\newblock Time series data imputation: A survey on deep learning approaches.
\newblock {\em arXiv preprint arXiv:2011.11347}, 2020.

\bibitem[\protect\citeauthoryear{Fortuin \bgroup \em et al.\egroup
  }{2020}]{fortuin2020gpvae}
Vincent Fortuin, Dmitry Baranchuk, Gunnar Raetsch, and Stephan Mandt.
\newblock {GP-VAE}: Deep probabilistic time series imputation.
\newblock In {\em AISTATS}, 2020.

\bibitem[\protect\citeauthoryear{Gong \bgroup \em et al.\egroup
  }{2021}]{gong2021missing}
Yongshun Gong, Zhibin Li, Jian Zhang, Wei Liu, Yilong Yin, and Yu~Zheng.
\newblock Missing value imputation for multi-view urban statistical data via
  spatial correlation learning.
\newblock {\em IEEE Transactions on Knowledge and Data Engineering}, 2021.

\bibitem[\protect\citeauthoryear{Goswami \bgroup \em et al.\egroup
  }{2024}]{goswami2024moment}
Mononito Goswami, Konrad Szafer, Arjun Choudhry, Yifu Cai, Shuo Li, and Artur
  Dubrawski.
\newblock Moment: A family of open time-series foundation models.
\newblock In {\em ICML}, 2024.

\bibitem[\protect\citeauthoryear{Gu \bgroup \em et al.\egroup
  }{2022}]{gu2022efficiently}
Albert Gu, Karan Goel, and Christopher Re.
\newblock Efficiently modeling long sequences with structured state spaces.
\newblock In {\em ICLR}, 2022.

\bibitem[\protect\citeauthoryear{Hamza{\c{c}}ebi}{2008}]{hamzaccebi2008improving}
Co{\c{s}}kun Hamza{\c{c}}ebi.
\newblock Improving artificial neural networks’ performance in seasonal time
  series forecasting.
\newblock {\em Information Sciences}, 178(23):4550--4559, 2008.

\bibitem[\protect\citeauthoryear{Ibrahim \bgroup \em et al.\egroup
  }{2012}]{ibrahim2012missing}
Joseph~G Ibrahim, Haitao Chu, and Ming-Hui Chen.
\newblock Missing data in clinical studies: issues and methods.
\newblock {\em Journal of clinical oncology}, 2012.

\bibitem[\protect\citeauthoryear{Jin \bgroup \em et al.\egroup
  }{2023}]{jin2023large}
Ming Jin, Qingsong Wen, Yuxuan Liang, Chaoli Zhang, Siqiao Xue, Xue Wang, James
  Zhang, Yi~Wang, Haifeng Chen, Xiaoli Li, et~al.
\newblock Large models for time series and spatio-temporal data: A survey and
  outlook.
\newblock {\em arXiv preprint arXiv:2310.10196}, 2023.

\bibitem[\protect\citeauthoryear{Jing \bgroup \em et al.\egroup
  }{2024}]{jing2024causality}
Baoyu Jing, Dawei Zhou, Kan Ren, and Carl Yang.
\newblock Causality-aware spatiotemporal graph neural networks for
  spatiotemporal time series imputation.
\newblock In {\em CIKM}, 2024.

\bibitem[\protect\citeauthoryear{Khayati \bgroup \em et al.\egroup
  }{2020}]{khayati2020mind}
Mourad Khayati, Alberto Lerner, Zakhar Tymchenko, and Philippe
  Cudr{\'e}-Mauroux.
\newblock Mind the gap: an experimental evaluation of imputation of missing
  values techniques in time series.
\newblock In {\em VLDB}, 2020.

\bibitem[\protect\citeauthoryear{Kim and Chi}{2018}]{kim2018tbm}
Yeo~Jin Kim and Min Chi.
\newblock {Temporal Belief Memory}: Imputing missing data during rnn training.
\newblock In {\em IJCAI}, 2018.

\bibitem[\protect\citeauthoryear{Kim \bgroup \em et al.\egroup
  }{2023}]{kim2023probabilistic}
Seunghyun Kim, Hyunsu Kim, Eunggu Yun, Hwangrae Lee, Jaehun Lee, and Juho Lee.
\newblock Probabilistic imputation for time-series classification with missing
  data.
\newblock In {\em ICML}, 2023.

\bibitem[\protect\citeauthoryear{Li \bgroup \em et al.\egroup
  }{2023}]{li2023data}
Xiao Li, Huan Li, Harry Kai-Ho Chan, Hua Lu, and Christian~S Jensen.
\newblock Data imputation for sparse radio maps in indoor positioning.
\newblock In {\em ICDE}, 2023.

\bibitem[\protect\citeauthoryear{Liang \bgroup \em et al.\egroup
  }{2024a}]{liang2024higher}
Guojun Liang, Prayag Tiwari, S{\l}awomir Nowaczyk, and Stefan Byttner.
\newblock Higher-order spatio-temporal physics-incorporated graph neural
  network for multivariate time series imputation.
\newblock In {\em CIKM}, 2024.

\bibitem[\protect\citeauthoryear{Liang \bgroup \em et al.\egroup
  }{2024b}]{liang2024foundation}
Yuxuan Liang, Haomin Wen, Yuqi Nie, Yushan Jiang, Ming Jin, Dongjin Song,
  Shirui Pan, and Qingsong Wen.
\newblock Foundation models for time series analysis: A tutorial and survey.
\newblock In {\em KDD}, 2024.

\bibitem[\protect\citeauthoryear{Little and
  Rubin}{2019}]{little2019statistical}
Roderick~JA Little and Donald~B Rubin.
\newblock {\em Statistical analysis with missing data}, volume 793.
\newblock John Wiley \& Sons, 2019.

\bibitem[\protect\citeauthoryear{Liu \bgroup \em et al.\egroup
  }{2019}]{liu2019naomi}
Yukai Liu, Rose Yu, Stephan Zheng, Eric Zhan, and Yisong Yue.
\newblock Naomi: Non-autoregressive multiresolution sequence imputation.
\newblock In {\em NeurIPS}, 2019.

\bibitem[\protect\citeauthoryear{Liu \bgroup \em et al.\egroup
  }{2022}]{liu2022multivariate}
Shuai Liu, Xiucheng Li, Gao Cong, Yile Chen, and Yue Jiang.
\newblock Multivariate time-series imputation with disentangled temporal
  representations.
\newblock 2022.

\bibitem[\protect\citeauthoryear{Liu \bgroup \em et al.\egroup
  }{2023}]{liu2023pristi}
Mingzhe Liu, Han Huang, Hao Feng, Leilei Sun, Bowen Du, and Yanjie Fu.
\newblock Pristi: A conditional diffusion framework for spatiotemporal
  imputation.
\newblock {\em arXiv preprint arXiv:2302.09746}, 2023.

\bibitem[\protect\citeauthoryear{Liu \bgroup \em et al.\egroup
  }{2024}]{liutimer}
Yong Liu, Haoran Zhang, Chenyu Li, Xiangdong Huang, Jianmin Wang, and Mingsheng
  Long.
\newblock Timer: Generative pre-trained transformers are large time series
  models.
\newblock In {\em Forty-first International Conference on Machine Learning},
  2024.

\bibitem[\protect\citeauthoryear{L{\"o}ning \bgroup \em et al.\egroup
  }{2019}]{loning2019sktime}
Markus L{\"o}ning, Anthony Bagnall, Sajaysurya Ganesh, Viktor Kazakov, Jason
  Lines, et~al.
\newblock sktime: A unified interface for machine learning with time series.
\newblock {\em arXiv preprint arXiv:1909.07872}, 2019.

\bibitem[\protect\citeauthoryear{Lugmayr \bgroup \em et al.\egroup
  }{2022}]{Lugmayr2022repaint}
Andreas Lugmayr, Martin Danelljan, Andres Romero, Fisher Yu, Radu Timofte, and
  Luc Van~Gool.
\newblock Repaint: Inpainting using denoising diffusion probabilistic models.
\newblock In {\em CVPR}, 2022.

\bibitem[\protect\citeauthoryear{Luo \bgroup \em et al.\egroup
  }{2018}]{luo2018grui}
Yonghong Luo, Xiangrui Cai, Ying ZHANG, Jun Xu, and Yuan xiaojie.
\newblock Multivariate time series imputation with generative adversarial
  networks.
\newblock In {\em NeurIPS}, 2018.

\bibitem[\protect\citeauthoryear{Luo \bgroup \em et al.\egroup
  }{2019}]{luo2019e2gan}
Yonghong Luo, Ying Zhang, Xiangrui Cai, and Xiaojie Yuan.
\newblock {E$^2$GAN}: End-to-end generative adversarial network for
  multivariate time series imputation.
\newblock In {\em IJCAI}, 2019.

\bibitem[\protect\citeauthoryear{Marisca \bgroup \em et al.\egroup
  }{2022}]{marisca2022learning}
Ivan Marisca, Andrea Cini, and Cesare Alippi.
\newblock Learning to reconstruct missing data from spatiotemporal graphs with
  sparse observations.
\newblock {\em NeurIPS}, 2022.

\bibitem[\protect\citeauthoryear{Miao \bgroup \em et al.\egroup
  }{2021}]{miao2021ssgan}
Xiaoye Miao, Yangyang Wu, Jun Wang, Yunjun Gao, Xudong Mao, and Jianwei Yin.
\newblock Generative semi-supervised learning for multivariate time series
  imputation.
\newblock In {\em AAAI}, 2021.

\bibitem[\protect\citeauthoryear{Miyaguchi \bgroup \em et al.\egroup
  }{2021}]{miyaguchi2021variational}
Kohei Miyaguchi, Takayuki Katsuki, Akira Koseki, and Toshiya Iwamori.
\newblock Variational inference for discriminative learning with generative
  modeling of feature incompletion.
\newblock In {\em ICLR}, 2021.

\bibitem[\protect\citeauthoryear{Moritz and
  Bartz-Beielstein}{2017}]{moritz2017imputets}
Steffen Moritz and Thomas Bartz-Beielstein.
\newblock {imputeTS: Time Series Missing Value Imputation in R}.
\newblock {\em {The R Journal}}, 2017.

\bibitem[\protect\citeauthoryear{Mulyadi \bgroup \em et al.\egroup
  }{2021}]{mulyadi2021uncertainty}
Ahmad~Wisnu Mulyadi, Eunji Jun, and Heung-Il Suk.
\newblock Uncertainty-aware variational-recurrent imputation network for
  clinical time series.
\newblock {\em IEEE Transactions on Cybernetics}, 52(9):9684--9694, 2021.

\bibitem[\protect\citeauthoryear{Nakagawa}{2015}]{nakagawa2015missing}
Shinichi Nakagawa.
\newblock Missing data: mechanisms, methods and messages.
\newblock pages 81--105. Oxford University Press Oxford, UK, 2015.

\bibitem[\protect\citeauthoryear{Nie \bgroup \em et al.\egroup
  }{2024}]{nie2024imputeformer}
Tong Nie, Guoyang Qin, Wei Ma, Yuewen Mei, and Jian Sun.
\newblock Imputeformer: Low rankness-induced transformers for generalizable
  spatiotemporal imputation.
\newblock In {\em KDD}, 2024.

\bibitem[\protect\citeauthoryear{Qian \bgroup \em et al.\egroup
  }{2025}]{qian2025deep}
Linglong Qian, Hugh~Logan Ellis, Tao Wang, Jun Wang, Robin Mitra, Richard
  Dobson, and Zina Ibrahim.
\newblock How deep is your guess? a fresh perspective on deep learning for
  medical time-series imputation.
\newblock {\em IEEE Journal of Biomedical and Health Informatics}, 2025.

\bibitem[\protect\citeauthoryear{Rubin}{1976}]{rubin1976missing}
Donald~B. Rubin.
\newblock {Inference and missing data}.
\newblock {\em Biometrika}, 63(3):581--592, 12 1976.

\bibitem[\protect\citeauthoryear{Silva \bgroup \em et al.\egroup
  }{2012}]{silva2012physionet}
Ikaro Silva, George Moody, Daniel~J Scott, Leo~A Celi, and Roger~G Mark.
\newblock Predicting in-hospital mortality of icu patients: The
  physionet/computing in cardiology challenge 2012.
\newblock {\em Computing in cardiology}, 39:245, 2012.

\bibitem[\protect\citeauthoryear{Tashiro \bgroup \em et al.\egroup
  }{2021}]{tashiro2021csdi}
Yusuke Tashiro, Jiaming Song, Yang Song, and Stefano Ermon.
\newblock {CSDI}: Conditional score-based diffusion models for probabilistic
  time series imputation.
\newblock In {\em NeurIPS}, 2021.

\bibitem[\protect\citeauthoryear{Van~Buuren and
  Groothuis-Oudshoorn}{2011}]{vanbuuren2011mice}
Stef Van~Buuren and Karin Groothuis-Oudshoorn.
\newblock mice: Multivariate imputation by chained equations in r.
\newblock {\em Journal of statistical software}, 45:1--67, 2011.

\bibitem[\protect\citeauthoryear{Vaswani \bgroup \em et al.\egroup
  }{2017}]{vaswani2017attention}
Ashish Vaswani, Noam Shazeer, Niki Parmar, Jakob Uszkoreit, Llion Jones,
  Aidan~N Gomez, et~al.
\newblock Attention is all you need.
\newblock In {\em NeurIPS}, 2017.

\bibitem[\protect\citeauthoryear{Wang \bgroup \em et al.\egroup
  }{2023}]{wang2023observed}
Xu~Wang, Hongbo Zhang, Pengkun Wang, Yudong Zhang, Binwu Wang, Zhengyang Zhou,
  and Yang Wang.
\newblock An observed value consistent diffusion model for imputing missing
  values in multivariate time series.
\newblock In {\em SIGKDD}, 2023.

\bibitem[\protect\citeauthoryear{Wang \bgroup \em et al.\egroup
  }{2024}]{wang2024timemixer++}
Shiyu Wang, Jiawei Li, Xiaoming Shi, Zhou Ye, Baichuan Mo, Wenze Lin, Shengtong
  Ju, Zhixuan Chu, and Ming Jin.
\newblock Timemixer++: A general time series pattern machine for universal
  predictive analysis.
\newblock {\em arXiv preprint arXiv:2410.16032}, 2024.

\bibitem[\protect\citeauthoryear{Wen \bgroup \em et al.\egroup
  }{2023}]{wen2023transformers}
Qingsong Wen, Tian Zhou, Chaoli Zhang, Weiqi Chen, Ziqing Ma, Junchi Yan, and
  Liang Sun.
\newblock Transformers in time series: A survey.
\newblock In {\em International Joint Conference on Artificial
  Intelligence(IJCAI)}, 2023.

\bibitem[\protect\citeauthoryear{Wu \bgroup \em et al.\egroup
  }{2023a}]{wu2023timesnet}
Haixu Wu, Tengge Hu, Yong Liu, Hang Zhou, Jianmin Wang, and Mingsheng Long.
\newblock {TimesNet: Temporal 2D-Variation Modeling for General Time Series
  Analysis}.
\newblock In {\em ICLR}, 2023.

\bibitem[\protect\citeauthoryear{Wu \bgroup \em et al.\egroup
  }{2023b}]{wu2023differentiable}
Yangyang Wu, Jun Wang, Xiaoye Miao, Wenjia Wang, and Jianwei Yin.
\newblock Differentiable and scalable generative adversarial models for data
  imputation.
\newblock {\em IEEE Transactions on Knowledge and Data Engineering}, 2023.

\bibitem[\protect\citeauthoryear{Yang \bgroup \em et al.\egroup
  }{2024a}]{yang2024frequency}
Xinyu Yang, Yu~Sun, Xiaojie Yuan, and Xinyang Chen.
\newblock Frequency-aware generative models for multivariate time series
  imputation.
\newblock In {\em The Thirty-eighth Annual Conference on Neural Information
  Processing Systems}, 2024.

\bibitem[\protect\citeauthoryear{Yang \bgroup \em et al.\egroup
  }{2024b}]{yang2024survey}
Yiyuan Yang, Ming Jin, Haomin Wen, Chaoli Zhang, Yuxuan Liang, Lintao Ma,
  Yi~Wang, Chenghao Liu, Bin Yang, Zenglin Xu, et~al.
\newblock A survey on diffusion models for time series and spatio-temporal
  data.
\newblock {\em arXiv preprint arXiv:2404.18886}, 2024.

\bibitem[\protect\citeauthoryear{Yoon \bgroup \em et al.\egroup
  }{2019}]{yoon2017mrnn}
Jinsung Yoon, William~R. Zame, and Mihaela van~der Schaar.
\newblock Estimating missing data in temporal data streams using
  multi-directional recurrent neural networks.
\newblock {\em IEEE Trans. on Biomedical Engineering}, 2019.

\bibitem[\protect\citeauthoryear{Yuan and Qiao}{2024}]{yuan2024diffusion}
Xinyu Yuan and Yan Qiao.
\newblock Diffusion-ts: Interpretable diffusion for general time series
  generation.
\newblock {\em arXiv preprint arXiv:2403.01742}, 2024.

\bibitem[\protect\citeauthoryear{Zhou \bgroup \em et al.\egroup
  }{2023}]{zhou2023one}
Tian Zhou, Peisong Niu, Liang Sun, Rong Jin, et~al.
\newblock One fits all: Power general time series analysis by pretrained lm.
\newblock {\em Advances in neural information processing systems},
  36:43322--43355, 2023.

\bibitem[\protect\citeauthoryear{Zhou \bgroup \em et al.\egroup
  }{2024}]{zhou2024mtsci}
Jianping Zhou, Junhao Li, Guanjie Zheng, Xinbing Wang, and Chenghu Zhou.
\newblock Mtsci: A conditional diffusion model for multivariate time series
  consistent imputation.
\newblock In {\em CIKM}, pages 3474--3483, 2024.

\end{thebibliography}

\end{document}